\documentclass[10pt,twocolumn,letterpaper]{article}

\usepackage[pagenumbers]{cvpr}

\def\confName{CVPR}
\def\confYear{2027}

\newif\ifincludesupp
\includesupptrue

\usepackage[T1]{fontenc}
\usepackage[utf8]{inputenc}
\usepackage{amsfonts}
\usepackage{amsthm}
\usepackage{tikz}
\usetikzlibrary{arrows.meta,positioning,fit,backgrounds,calc,shapes.geometric}

\definecolor{pcblue}{HTML}{1F77B4}
\definecolor{pcgreen}{HTML}{2CA02C}
\definecolor{pcred}{HTML}{D62728}
\definecolor{pcorange}{HTML}{FF7F0E}
\definecolor{pcpurple}{HTML}{7B5BA6}
\definecolor{pcgray}{HTML}{6E6E6E}

\newtheorem{observation}{Observation}[section]

\definecolor{cvprblue}{rgb}{0.21,0.49,0.74}
\usepackage[pagebackref,breaklinks,colorlinks,allcolors=cvprblue]{hyperref}
\usepackage{orcidlink}   % must follow hyperref

\title{PixCon: Clean-Positive Contrastive Learning\\
       for Foundation-Model Semi-Supervised Segmentation}

\author{
  Ebenezer Tarubinga~\orcidlink{0009-0004-7340-1873}\thanks{Code: \texttt{github.com/psychofict/PixCon}.}\\
  Ebenworks Systems, Seoul, Korea\\
  \texttt{research@ebenworks.co}
}

\begin{document}

\maketitle

\begin{abstract}
Can a simple labeled-pixel contrastive branch improve semi-supervised segmentation with a
strong pretrained encoder? We study this question with PixCon, which adds a class-wise memory
bank to a DINOv2-Base implementation of UniMatch~V2. The bank stores ground-truth-labeled
features only where the student prediction agrees with the label. A multi-positive contrastive
loss uses these features to train the shared encoder--decoder; the head and bank are discarded
at inference. This simplifies earlier labeled-pixel banks by removing confidence thresholds and
learned ranking. On Pascal~VOC with 183 labels, PixCon improves all three paired seeds by
$0.20$, $0.64$ and $1.83$ mIoU points, raising the mean from $87.01$ to $87.90$.
With 92 labels the mean gain is $0.40$ points; single-seed differences on Cityscapes and ADE20K
range from $-0.08$ to $+0.14$ points. A predicted-label contrastive control reaches nearly the
same Pascal accuracy, so these results support the added branch rather than an accuracy benefit
from its correctness filter. An exact gradient decomposition distinguishes mislabeled-positive
counts from their similarity-weighted influence. Separate teacher diagnostics find retained-pixel
error rates of $1.8\%$ on Pascal and $10.6\%$ on ADE20K. These diagnostics do not measure the
training bank or predict final accuracy. PixCon adds $15.3$--$19.7\%$ iteration time in the
measured runs and no inference computation. We report paired seeds, training cost and the limits
of the evidence.

\end{abstract}

\begin{figure}[t]
\centering
\includegraphics[width=\linewidth]{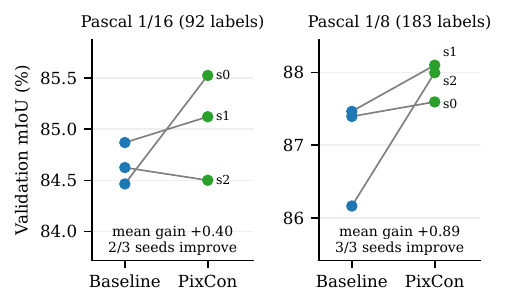}
\caption{\textbf{Paired outcomes on Pascal VOC.} Best validation EMA-teacher mIoU for
our UniMatch~V2 implementation and PixCon, with the same labeled split and training seed in
each pair. All three seeds improve at 1/8; two improve at 1/16. The smaller observed standard
deviation at 1/8 is descriptive, based on three runs. Table~\ref{tab:pascal} gives means and
sample standard deviations.}

\label{fig:teaser}
\end{figure}

\section{Introduction}
\label{sec:intro}

Semantic segmentation assigns a class to every pixel. Dense annotations are expensive:
a Cityscapes image takes about 1.5 hours to annotate~\cite{cityscapes}. Semi-supervised semantic
segmentation (SSSS) combines a small labeled set with unlabeled images. A common approach
trains a student to reproduce confident teacher predictions under stronger image
augmentations~\cite{fixmatch,unimatch,unimatchv2}. The teacher is an exponential moving average
(EMA) of the student weights.

Strong pretrained encoders change the baseline against which an auxiliary objective must be
judged. UniMatch~V2~\cite{unimatchv2} uses DINOv2~\cite{dinov2} and reports $87.9$ mIoU on
Pascal VOC with 183 labeled images. Improvements over earlier ResNet systems cannot establish
that an additional training component helps this stronger baseline. We therefore hold the
encoder, data split, optimizer and consistency objective fixed, and study the addition of a
labeled-pixel contrastive branch.

\paragraph{A simple branch with a specific scope.}
PixCon stores pixel embeddings in one queue per class. Only labeled pixels whose student
prediction agrees with the ground truth enter the queues and act as contrastive anchors.
The loss increases the total similarity to same-class entries relative to the whole bank.
The branch trains the shared encoder and decoder, but adds no inference computation.
Labeled-pixel memory banks are established prior work: Alonso~et~al.~\cite{alonso2021} select
correct, confident teacher features and rank them with learned attention. PixCon uses student
features, removes the confidence threshold and ranking network, and applies a positive-and-negative
contrastive objective to labeled anchors only. Its contribution is this simple configuration
and an evaluation of its behavior alongside a foundation-model consistency baseline.

\paragraph{Separate the branch from the filter.}
Across three seeds on one Pascal 1/8 split, the branch increases mean mIoU by $0.89$ points.
All three paired differences are positive, with the largest gain on the weakest baseline run
(Fig.~\ref{fig:teaser}). At 1/16 the mean gain is $0.40$ points, with one regression.
The single-seed comparisons on Pascal 1/4, Cityscapes and ADE20K have small differences.
These observations motivate a narrower question: does the correctness filter explain the gain?
Our single-seed control changes the contrastive selection and label assignment to confident
student predictions on the same labeled images. Its Pascal scores differ from PixCon by at most
$0.02$ points. Thus, the available comparison does not establish a distinct accuracy benefit
from clean admission. It supports studying the added contrastive branch as a whole.

\paragraph{What cleanliness guarantees.}
Ground-truth labels prevent pseudo-label assignment errors in the bank, relative to the dataset
annotations. This property follows from labeled membership alone; the additional correctness
gate selects features the current student classifies correctly. Neither condition guarantees
correct annotation, useful feature geometry or improved generalization. We derive the exact
per-anchor gradient to distinguish the fraction of mislabeled positives from their
similarity-weighted contribution. A separate diagnostic measures the error among confident
EMA-teacher predictions on validation data. It gives $1.8\%$ on Pascal and $10.6\%$ on ADE20K,
but cannot identify the contamination of a per-class student bank during training.

Our contributions are:
\begin{enumerate}
\item A labeled-pixel contrastive branch with a correctness gate, class-wise queues and no
bank confidence threshold or ranking network, evaluated with DINOv2-Base.
\item A paired comparison with the same consistency implementation, reporting all seeds,
cross-dataset outcomes, per-image changes and measured training overhead.
\item A distinction between annotation-consistent bank labels, similarity-weighted gradient
influence and measured accuracy, supported by an exact decomposition and a predicted-label control.
\end{enumerate}

\section{Related Work}
\label{sec:related}

\paragraph{Consistency and foundation encoders.}
Mean Teacher~\cite{meanteacher}, CPS~\cite{cps}, ST++~\cite{stplusplus},
PS-MT~\cite{psmt}, AugSeg~\cite{augseg} and UniMatch~\cite{unimatch} develop ways to learn from
unlabeled images through predictions and perturbations. UniMatch~V2~\cite{unimatchv2} combines
DINOv2 features with two strongly augmented views and complementary feature dropout. We adopt
that design in a local implementation; section~\ref{sec:setup_exp} specifies its differences
from the published protocol. DINOv2 pretraining is held fixed across our comparisons, so the
reported paired differences concern the added branch, not the value of foundation pretraining.

\paragraph{Pixel contrastive learning.}
The closest antecedent is the class-wise labeled memory bank of Alonso~et~al.~\cite{alonso2021}.
Its teacher features must be correctly classified and confident, and a learned module ranks
stored entries. Student features from labeled and unlabeled images are aligned with same-class
bank features. PixCon retains the labeled-and-correct rule, uses student embeddings, omits the
ranking and bank confidence threshold, and uses labeled anchors with other-class negatives.
We do not claim the first clean bank or an empirical advantage over Alonso~et~al.; no matched
reproduction of that method is included.

ReCo~\cite{reco} forms class representations using labeled and confidently pseudo-labeled pixels
and samples hard negatives. U\textsuperscript{2}PL~\cite{u2pl} uses uncertain pixels as negatives
and constructs class centers from reliable pixels. Their sampling, representations and loss
construction differ from PixCon. Our predicted-label control is a variant of our own branch,
not an implementation of either method. The loss is related to supervised contrastive
learning~\cite{supcon}, but places the sum over positive exponentials inside the logarithm;
this distinction matters for the gradient in section~\ref{sec:cleanpos}.

\paragraph{Alternative uses of features and uncertainty.}
AllSpark~\cite{allspark} exchanges labeled and unlabeled features; CorrMatch~\cite{corrMatch}
propagates labels through correlations; SemiVL~\cite{semivl} adds vision-language guidance;
CW-BASS~\cite{cwbass} weights pseudo-labels by confidence and adds boundary-aware learning.
SWSEG~\cite{swseg} studies feature alignment and uniformity, and
Li~et~al.~\cite{encodingperspective} study pseudo-label learning from an encoding perspective.
These methods motivate examining representations beyond output consistency, but their reported
results are not controlled comparisons with our implementation.
Recent work also improves the reliability or optimization of foundation-model SSSS:
POS~\cite{pos2025} balances supervised and unsupervised gradients, while
HESS~\cite{hess2026} uses evidential uncertainty and reports gains with DINOv2-Small.
Confidence Separable Learning~\cite{csl} revisits selection, and BoundMatch~\cite{boundmatch}
adds boundary supervision. These results preclude a general claim that pseudo-label selection
ceases to matter with foundation models. PixCon instead keeps one selection rule fixed and
examines a separate labeled-data objective.

\paragraph{Noisy contrastive targets.}
RINCE~\cite{rince} studies contrastive learning with noisy views, and
Gerard and Sullivan~\cite{gerard2022noisy} study noisy positive pairs in segmentation
pretraining. Our analysis concerns the positive-set loss actually used by PixCon. It shows why
an unweighted error rate alone cannot determine the contrastive update or its effect on
segmentation accuracy.

\begin{figure*}[t]
\centering
\includegraphics[width=\textwidth]{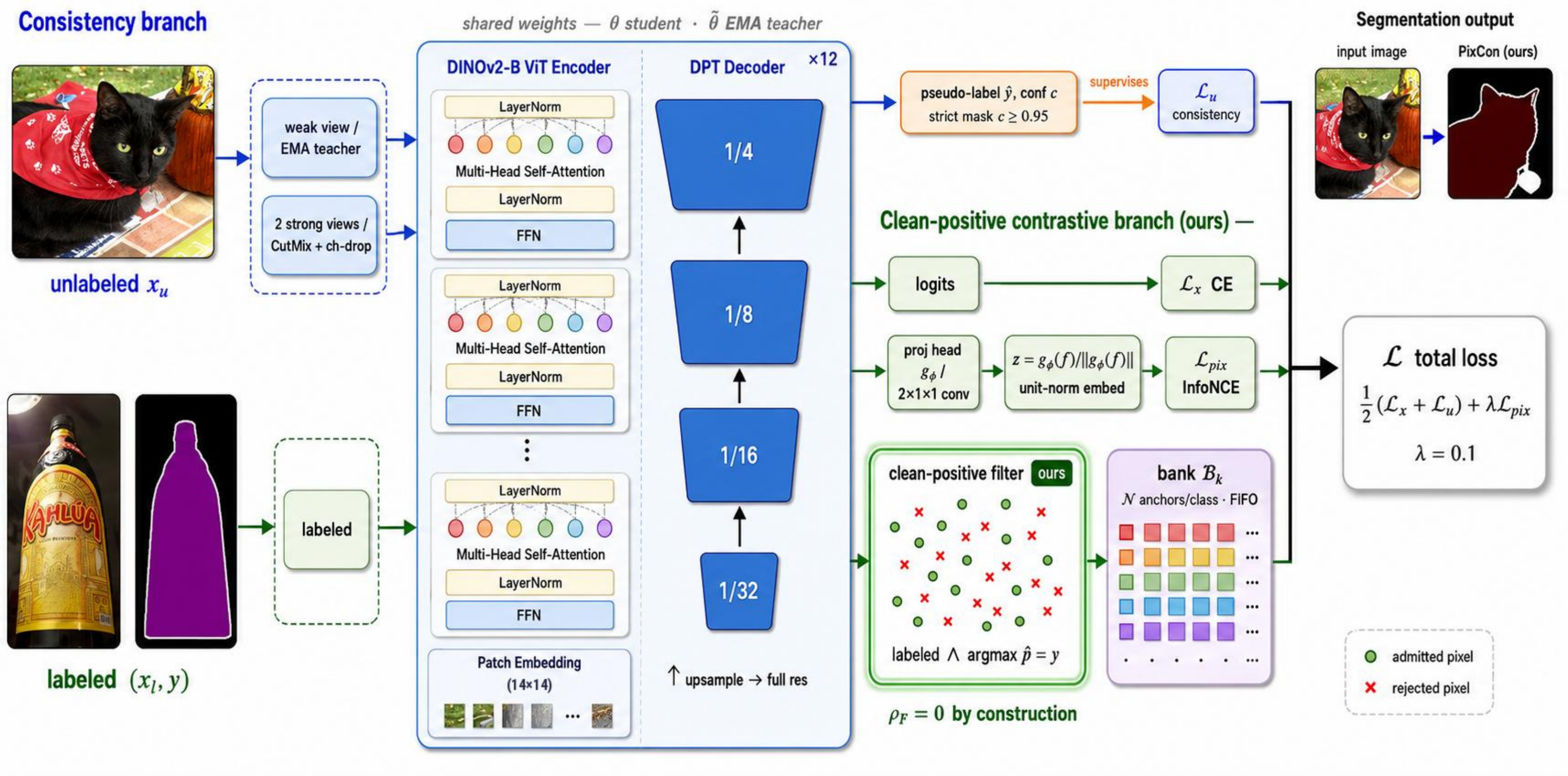}
\caption{\textbf{Training with PixCon.} The student uses one DINOv2-B encoder and DPT-style
decoder for all image views. The EMA teacher supplies confidence-masked targets for two strongly
augmented unlabeled views. On labeled images, the projection head produces normalized embeddings;
the ground-truth label and student prediction determine eligible anchors. The contrastive loss
reads the previous bank, then detached anchors enter their class queues. Supervised and
consistency losses also update the shared student. At inference only the segmentor is used.}
\label{fig:method_overview}
\end{figure*}

\section{Method}
\label{sec:method}

PixCon adds a training-only contrastive objective to an EMA teacher--student segmentation
system (Fig.~\ref{fig:method_overview}). The student shares its encoder and decoder between
supervised segmentation, unlabeled consistency and the contrastive branch. Disabling the branch
recovers our consistency baseline.

\subsection{Problem Setup}
\label{sec:setup}

Given a labeled set $\mathcal{D}_l = \{(x_i, y_i)\}_{i=1}^{N_l}$ and unlabeled set
$\mathcal{D}_u = \{x_j\}_{j=1}^{N_u}$ ($N_u \gg N_l$), a teacher $f_{\bar\theta}$ (EMA of the
student) produces predictions $p(x) \in \mathbb{R}^{K \times H \times W}$, pseudo-labels
$\hat{y} = \arg\max_k p_k(x)$, and confidences $c = \max_k p_k(x)$, filtered by a global
threshold into $\mathcal{M} = \{(h,w) : c_{h,w} \geq \tau\}$. The consistency branch uses a strict
$\tau{=}0.95$ and supervises the student on two strong-augmented views with complementary channel
dropout (section~\ref{sec:integration}). In parallel the \emph{clean-positive contrastive branch}
(section~\ref{sec:pixcon}) shapes the shared embedding space from the labeled set; both are trained
jointly over one encoder--decoder.

\subsection{The Clean-Positive Contrastive Branch}
\label{sec:pixcon}

A small projection head $g_\phi$ maps the fused decoder feature
$f \in \mathbb{R}^{C \times H' \times W'}$ to a normalized embedding space:
\begin{equation}
    z = \frac{g_\phi(f)}{\lVert g_\phi(f) \rVert_2}
      \in \mathbb{R}^{D \times H' \times W'},
    \qquad D = 256.
\end{equation}
$g_\phi$ is two $1{\times}1$ convolutions with batch-norm and ReLU in between; its
parameters join the optimizer at the decoder learning rate.

\paragraph{The clean-positive bank.}
We maintain a per-class FIFO queue
$\mathcal{B} = \{\mathcal{B}_k\}_{k=1}^K$ of unit-norm pixel embeddings, with at most $N{=}256$
entries per class. Each labeled image's decoder feature is projected to $z$ and we select
\emph{anchor pixels} satisfying two conditions:
\begin{equation}
    \begin{aligned}
    \mathcal{A} = \{(b, h, w) : \;& y_{b,h,w} \neq \mathrm{ignore} \\[-1pt]
                              &\land\;
                              \arg\max_k \mathrm{logits}_{b,k,h,w} = y_{b,h,w}\}.
    \end{aligned}
    \label{eq:clean-anchor}
\end{equation}
We sample at most $m{=}64$ eligible pixels per class. Classes with fewer eligible pixels
contribute all of them, so this is capped per-class sampling rather than equal class counts.
We first compute the loss against the previous bank, then enqueue all sampled anchors into
$\mathcal{B}_k$ with gradients detached. Thus, a current anchor is not its own positive.
The rule in \eqref{eq:clean-anchor} combines labeled membership with prediction--label agreement: a bank entry is added
only when the embedding is at a location the student already classifies correctly. The predicate never reads a pseudo-label,
so no pseudo-label assignment error can enter; and unlike the labeled bank of
Alonso~et~al.~\cite{alonso2021} it needs no confidence threshold and no ranking network.

\paragraph{Multi-positive contrastive loss.}
For an anchor $z_a$ with label $y_a$, let $\mathcal{P}_a \subseteq \mathcal{B}$ be all bank
entries with the same class and $\mathcal{N}_a$ the rest. The contrastive loss is the
positive-set log-ratio form, related to supervised contrastive learning~\cite{supcon}:
\begin{equation}
    \mathcal{L}_{\mathrm{pix}}
    = -\frac{1}{|\mathcal{A}_+|}\!\!\sum_{a \in \mathcal{A}_+}\!\!
        \log
        \frac{\sum_{p \in \mathcal{P}_a}\exp(z_a\!\cdot\!p/\eta)}
             {\sum_{n \in \mathcal{P}_a \cup \mathcal{N}_a}\exp(z_a\!\cdot\!n/\eta)},
    \label{eq:pixcon-loss}
\end{equation}
where $\eta{=}0.1$ is the temperature and $\mathcal{A}_+$ is the subset of anchors with at
least one same-class bank entry. We cap $|\mathcal{A}_+|$ at 1024 per iteration to bound
compute. Each retained anchor is compared with all stored entries, at most $KN$, giving
$O(|\mathcal{A}_+|KND)$ arithmetic for the dot products. The global cap is sampled uniformly
before discarding anchors without positives. This loss uses a log of summed positive scores;
it is not the average of per-positive log losses commonly used in SupCon.

\paragraph{Total objective.}
PixCon's two branches are trained jointly under a single objective, with the contrastive
term weighted by $\lambda_{\mathrm{pix}}$:
\begin{equation}
    \mathcal{L} = \tfrac{1}{2}\bigl(\mathcal{L}_x + \mathcal{L}_u\bigr)
                  + \lambda_{\mathrm{pix}}\,\mathcal{L}_{\mathrm{pix}},
    \qquad \lambda_{\mathrm{pix}} = 0.1,
\end{equation}
where $\mathcal{L}_x$ is the supervised cross-entropy on labeled pixels and $\mathcal{L}_u$
is the dual-stream consistency loss of section~\ref{sec:integration}. The bank starts empty, so the first loss is zero. A class contributes to the contrastive
loss once it has a same-class entry; a partially filled queue is sufficient. The supervised
and consistency objectives remain active for every class. Stored embeddings are detached
snapshots and are not recomputed when the model changes. The cleanliness claim concerns their
assigned labels, not whether older features remain correctly classified by the current model.

\subsection{What Positive-Label Errors Change}
\label{sec:cleanpos}

For one anchor, let $P$ be the bank entries assigned to its class and $B$ the full bank.
The per-anchor loss in Eq.~\eqref{eq:pixcon-loss} is
$\ell_a=-\log\sum_{p\in P}e^{z_a^\top p/\eta}
+\log\sum_{b\in B}e^{z_a^\top b/\eta}$.
Define positive-normalized weights $q_p=e^{z_a^\top p/\eta}/\sum_{u\in P}e^{z_a^\top u/\eta}$
and full-bank weights $r_b=e^{z_a^\top b/\eta}/\sum_{u\in B}e^{z_a^\top u/\eta}$.
With detached bank entries, differentiation gives
\begin{equation}
 -\eta\nabla_{z_a}\ell_a
 = \mu_P-\mu_B,\quad
 \mu_P=\sum_{p\in P}q_pp,\quad \mu_B=\sum_{b\in B}r_bb.
 \label{eq:exact-gradient}
\end{equation}
This is the gradient with respect to the normalized embedding; backpropagation through
normalization also projects it onto the tangent space of the unit sphere.

\begin{observation}[Count contamination and softmax mass]
Assume\label{prop:gradient} the anchor's class is correct. Partition $P$ into entries with correct labels $T$
and incorrect labels $F$. Let $\alpha=\sum_{p\in F}q_p$, and let $\mu_T,\mu_F$ be the
separately normalized softmax means of the two subsets. Then
\begin{equation}
 -\eta\nabla_{z_a}\ell_a
 =(1-\alpha)\mu_T+\alpha\mu_F-\mu_B.
 \label{eq:mixture-gradient}
\end{equation}
For unit-norm entries and nonempty $T$,
$\|\mu_P-\mu_T\|\leq2\alpha$.
Writing $\rho_F=|F|/|P|$ and
$\bar e_S=|S|^{-1}\sum_{p\in S}\exp(z_a^\top p/\eta)$ gives, when both subsets are nonempty,
\begin{equation}
 \frac{\alpha}{1-\alpha}
 =\frac{\rho_F}{1-\rho_F}\frac{\bar e_F}{\bar e_T}.
 \label{eq:mass-ratio}
\end{equation}
Thus, count contamination alone does not determine the false-positive softmax mass.
\end{observation}

The coefficient ratio reduces to $\rho_F/(1-\rho_F)$ only if the mean exponentiated
similarities are equal. It is not a ratio of gradient norms: vector cancellation and the
full-bank term also matter. The bound holds for the change in the positive mean with
fixed embeddings; replacing entries can change the denominator as well. No final-accuracy
prediction follows from this calculation. The supplement gives the derivation and scope.

\paragraph{Annotation-consistent positives.}
For a ground-truth-labeled anchor and bank, $F$ is empty relative to those annotations and
$\alpha=\rho_F=0$. Any bank that uses ground-truth class assignments has this property,
including Alonso~et~al.'s labeled bank~\cite{alonso2021}. PixCon's correctness gate further
selects features by the current student prediction; it is not needed for the label-source
property. Zero assignment error does not eliminate negative-class repulsion, annotation noise,
stale embeddings or uncertainty in the final model.

\paragraph{A separate teacher diagnostic.}
On validation images, we measure the fraction of erroneous EMA-teacher predictions among
non-void pixels with confidence at least 0.95. Denote it $e_{\mathrm{ret}}$ to distinguish
it from the class- and anchor-dependent bank quantity $\rho_F$. It is $0.018$ on Pascal 1/8
and $0.106$ on ADE20K 1/8. These pixel-pooled measurements on the baseline's best-EMA checkpoint describe the
teacher's retained predictions. They do not measure the student-based, per-class-sampled
training bank or its softmax mass $\alpha$. The control in section~\ref{sec:ablation}
evaluates the two contrastive labeling rules empirically, without substituting
$e_{\mathrm{ret}}$ into a prediction of their accuracy difference.

\subsection{Architecture, Training Loop and Cost}
\label{sec:integration}

We use a DINOv2-Base ViT-B/14 encoder~\cite{dinov2} with a lightweight DPT-style~\cite{dpt}
fusion decoder that combines four intermediate transformer layers into a per-pixel logit map, and
use a UniMatch~V2-style fine-tuning protocol (hyperparameters in
section~\ref{sec:setup_exp}).

\paragraph{Dual perturbation consistency.}
The teacher predicts a class and confidence on each weak unlabeled view. Independent strong
augmentation and CutMix produce two student views and their corresponding mixed targets.
A concatenated forward pass applies complementary channel masks to their fused features.
For each stream $j\in\{1,2\}$, let $V_j$ be its valid pixels. We use
\begin{equation}
\mathcal{L}_{u,j}=\frac{1}{|V_j|}\sum_{i\in V_j}
\mathbf{1}[c_{j,i}\geq0.95]\,\mathrm{CE}(p_{j,i},\hat y_{j,i}),
\end{equation}
and $\mathcal{L}_u=(\mathcal{L}_{u,1}+\mathcal{L}_{u,2})/2$. The denominator counts all valid
pixels, including those rejected by the mask. These are plain masked cross-entropies, without
an additional boundary loss or continuous confidence weighting. The teacher decay is
$\gamma_i=\min(1-1/(i+1),0.996)$ at zero-based iteration $i$.

\paragraph{Cost.}
The branch adds a projection head of two $1{\times}1$ convolutions ($256{\to}256{\to}256$,
$131{,}840$ parameters, $0.13\%$ of the segmentor), a bank of $K{\cdot}N{\cdot}D$ floats
($5.5$~MB on Pascal, $5.0$~MB on Cityscapes, $39.3$~MB on ADE20K in fp32), one head forward over
the labeled batch and an InfoNCE over at most $1024$ anchors per iteration. Measured on paired
runs in the same container, a PixCon iteration takes $15\%$ longer than a UniMatch~V2 iteration on
Pascal~1/4 and $19$--$20\%$ longer on Cityscapes (supplement). At inference the head and bank are
discarded; the deployed architecture matches our consistency baseline.

\section{Experiments}
\label{sec:experiments}

\subsection{Setup}
\label{sec:setup_exp}

\paragraph{Datasets and splits.}
We use PASCAL VOC 2012~\cite{pascalvoc} with the augmented training pool~\cite{sbd}
(10,582 train, 1,449 validation images, 21 classes), Cityscapes~\cite{cityscapes}
(2,975/500, 19 classes) and ADE20K~\cite{ade20k} (20,210/2,000, 150 classes).
Pascal labels are sampled from its 1,464 high-quality training annotations: 92, 183 or 366
images. All remaining training images are unlabeled. Cityscapes uses 186 or 372 labels;
ADE20K uses 2,526. The labeled file list is fixed across seeds in each cell. Both methods
use seeds 0--2 on Pascal 1/16 and 1/8; other cells use seed 0. Thus, the reported variation
is across training seeds, not independently sampled label splits.

\paragraph{Training and evaluation.}
The encoder is DINOv2-Base ViT-B/14. AdamW uses learning rates $5\times10^{-6}$ for the
encoder and $2\times10^{-4}$ for the decoder and projection head, weight decay $0.01$ and
polynomial decay with power $0.9$. Each iteration uses 16 labeled and 16 unlabeled images;
the labeled set is oversampled. The crop is $518\times518$, or $686\times686$ for Cityscapes.
The maximum schedule is 60 epochs, or 120 for Cityscapes, with early stopping after 20 epochs
without validation EMA improvement. We report each run's best validation EMA mIoU, averaged
over classes including Pascal background and excluding void pixels. Means and sample standard
deviations use all three seeds; differences are computed before rounding. Model selection and
reporting use the same validation set, so these are validation results, not independent test
estimates. Sensitivity sweeps also use validation performance.

Pascal and ADE20K evaluation use a single whole-image prediction after reflection padding to
multiples of 14. Cityscapes uses overlapping $686$-pixel windows with stride two-thirds of
the crop and averages their softmax probabilities. We use no multi-scale or flip evaluation.
The supplement specifies the architecture, augmentation, run provenance and post-hoc diagnostics.

\paragraph{What is controlled.}
Our baseline and PixCon use the same local implementation, split, seed, optimizer, maximum
schedule and stopping rule; the added branch changes training computation and random-number
consumption. They are protocol-matched, not equal-compute runs. We retain
$\lambda_\mathrm{pix}=0.1$, $\eta=0.1$, $D=N=256$, the per-class cap 64 and global anchor cap
1024 across datasets. Published UniMatch~V2 results use a different DPT implementation, different
Cityscapes settings (crop 798, 180 epochs and supervised OHEM), and a different image-resizing
policy at evaluation. Consequently, those rows are reference results; they are not our matched
baseline. Published rows for other methods are transcribed from UniMatch~V2's comparison tables
unless noted. There is no matched supervised-only or prior contrastive-method control.

\subsection{Main Results}
\label{sec:main_results}

\begin{table*}[t]
\centering
\caption{\textbf{Pascal VOC 2012, validation mIoU (\%).} Headers give labeled-image
counts. Published reference rows are quoted from UniMatch~V2~\cite{unimatchv2}; PrevMatch is
from its own paper~\cite{prevmatch}. Only the last two rows use our implementation.
Their 1/16 and 1/8 results are mean$\pm$sample std over three training seeds on one fixed
split; $\dagger$ denotes one seed. Bold marks the larger observed mean within our paired rows,
not statistical significance. ``--'' means not run.}
\label{tab:pascal}
\footnotesize
\setlength{\tabcolsep}{6pt}
\begin{tabular}{llccccc}
\toprule
Method & Encoder & 1/16\,(92) & 1/8\,(183) & 1/4\,(366) & 1/2\,(732) & Full\,(1464) \\
\midrule
\multicolumn{7}{l}{\emph{Specialized backbones (prior work):}} \\
Supervised baseline                  & RN-101  & 45.1 & 55.3 & 64.8 & 69.7 & 73.5 \\
ST++~\cite{stplusplus}               & RN-101  & 65.2 & 71.0 & 74.6 & 77.3 & 79.1 \\
U\textsuperscript{2}PL~\cite{u2pl}   & RN-101  & 68.0 & 69.2 & 73.7 & 76.2 & 79.5 \\
PS-MT~\cite{psmt}                    & RN-101  & 65.8 & 69.6 & 76.6 & 78.4 & 80.0 \\
AugSeg~\cite{augseg}                 & RN-101  & 71.1 & 75.5 & 78.8 & 80.3 & 81.4 \\
CW-BASS~\cite{cwbass}                & RN-50   & 72.8 & 75.8 & 76.2 & 77.2 & --   \\
UniMatch~\cite{unimatch}             & RN-101  & 75.2 & 77.2 & 78.8 & 79.9 & 81.2 \\
CorrMatch~\cite{corrMatch}           & RN-101  & 76.4 & 78.5 & 79.4 & 80.6 & 81.8 \\
DDFP~\cite{ddfp}                     & RN-101  & 75.0 & 78.0 & 79.5 & 81.2 & 82.0 \\
PrevMatch~\cite{prevmatch}           & RN-101  & 77.0 & 78.5 & 79.6 & 80.4 & 81.6 \\
BeyondPixels~\cite{beyondpixels}     & RN-101  & 77.3 & 78.6 & 79.8 & 80.8 & 81.7 \\
AllSpark~\cite{allspark}             & MiT-B5  & 76.1 & 78.4 & 79.8 & 80.8 & 82.1 \\
SemiVL~\cite{semivl}                 & CLIP-B  & 84.0 & 85.6 & 86.0 & 86.7 & 87.3 \\
\midrule
\multicolumn{7}{l}{\emph{DINOv2 reference results and our paired runs:}} \\
UniMatch~V2 (published)~\cite{unimatchv2} & DINOv2-S & 79.0 & 85.5 & 85.9 & 86.7 & 87.8 \\
UniMatch~V2 (published)~\cite{unimatchv2} & DINOv2-B & 86.3 & 87.9 & 88.9 & 90.0 & 90.8 \\
UniMatch~V2 (our repro)                  & DINOv2-B & $\mathit{84.65}{\scriptstyle\pm0.20}$ & $\mathit{87.01}{\scriptstyle\pm0.73}$ & $\mathit{88.59}^{\dagger}$ & -- & -- \\
PixCon (ours)                            & DINOv2-B & $\mathbf{85.05}{\scriptstyle\pm0.52}$ & $\mathbf{87.90}{\scriptstyle\pm0.27}$ & $\mathbf{88.68}^{\dagger}$ & -- & -- \\
\bottomrule
\end{tabular}
\end{table*}

\begin{figure}[t]
\centering
\includegraphics[width=\linewidth]{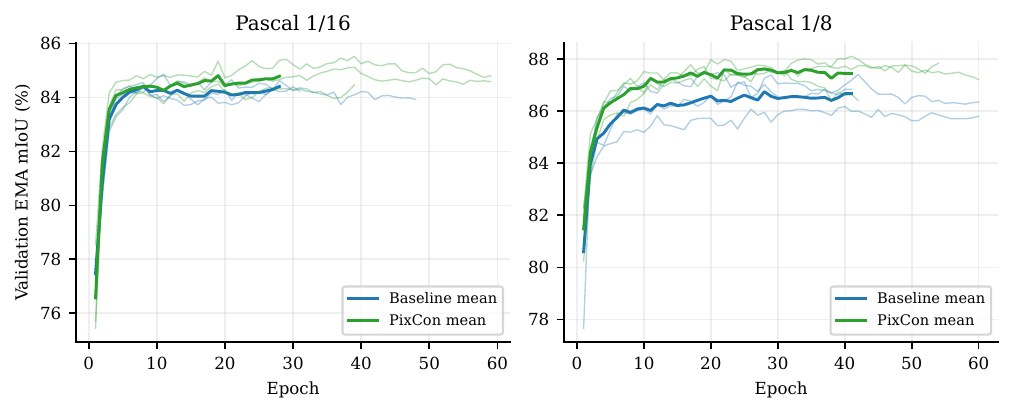}
\caption{\textbf{Training trajectories on fixed Pascal splits.} Best-checkpoint selection
uses the EMA validation curves. Thin lines show each seed; thick lines show the three-seed
mean only while all six runs are active. At 1/8 the PixCon mean is higher in 41/41 common
epochs; at 1/16 it is higher in 26/28, with one tie at the logged precision. Individual trajectories continue after early stopping
ends another run. Epoch-wise comparisons do not increase the number of independent seeds.}
\label{fig:curves}
\end{figure}

\paragraph{Paired Pascal results.}
At 1/8, the three baseline/PixCon pairs are $87.40/87.60$, $87.47/88.10$ and $86.17/88.00$
(Table~\ref{tab:pascal}). Computed from unrounded scores, the paired gains are $+0.20$, $+0.64$
and $+1.83$ points. Mean mIoU increases from $87.01\pm0.73$ to $87.90\pm0.27$.
The largest gain is on the weakest baseline seed; the observed standard deviation falls, but
three runs do not establish variance reduction as a general property. At 1/16, the paired gains
are $+1.06$, $+0.25$ and $-0.12$ points, for a mean gain of $0.40$.
The matched-seed mean curve is higher in all 41 common epochs at 1/8
(Fig.~\ref{fig:curves}); these correlated epochs are not independent replications.

\paragraph{Other datasets and budgets.}
The Pascal 1/4 difference is $+0.09$ points. Cityscapes differences are $+0.03$ at 1/16 and
$-0.08$ at 1/8 (Table~\ref{tab:cityscapes}); ADE20K differs by $+0.14$
(Table~\ref{tab:ade20k}). These are small single-seed differences, without measured uncertainty
for those cells. They establish neither an improvement nor statistical equivalence.
PixCon's Pascal 1/8 mean numerically matches the published UniMatch~V2-B result, but the
protocol differences prevent a performance-ranking claim against that published model.

\begin{table}[t]
\centering
\caption{\textbf{Single-seed outcomes beyond scarce-label Pascal.} Validation mIoU at
best EMA, seed 0. All rows compare our baseline and PixCon under the same dataset-specific
protocol. These small differences have no across-run uncertainty estimate. The supplement
retains published Cityscapes and ADE20K reference tables.}
\label{tab:cityscapes}
\label{tab:ade20k}
\small
\setlength{\tabcolsep}{4pt}
\begin{tabular}{lrccc}
\toprule
Dataset & labels & Baseline & PixCon & $\Delta$ \\
\midrule
Pascal & 366 & 88.59 & 88.68 & $+0.09$ \\
Cityscapes & 186 & 83.16 & 83.20 & $+0.03$ \\
Cityscapes & 372 & 83.96 & 83.88 & $-0.08$ \\
ADE20K & 2526 & 49.10 & 49.23 & $+0.14$ \\
\bottomrule
\end{tabular}
\end{table}

\subsection{Per-Image and Qualitative Comparison}
\label{sec:qualitative}

\begin{table*}[t]
\centering
\caption{\textbf{All validation images, not selected ones.} Per-image comparison of PixCon and
the UniMatch~V2 reproduction (seed~0, EMA teachers, whole-image inference). An image is a win when PixCon has fewer
wrongly labeled pixels; $p$ is a two-sided sign test over wins and losses; error is the pooled
per-pixel error rate (\%, lower is better). Seed~0 is the largest-gain seed at Pascal~1/16 and
the smallest at 1/8 (Fig.~\ref{fig:teaser}).}
\label{tab:perimage}
\small
\setlength{\tabcolsep}{8pt}
\begin{tabular}{lrrrrcc}
\toprule
Cell & images & win & tie & loss & $p$ & error UM2$\to$PixCon \\
\midrule
Pascal 1/16     & 1449 & 935  & 2  & 512 & ${<}10^{-6}$ & $3.92\to3.66$ \\
Pascal 1/8      & 1449 & 753  & 16 & 680 & $0.057$      & $2.76\to2.67$ \\
Cityscapes 1/16 & 500  & 242  & 0  & 258 & $0.50$       & $3.76\to3.77$ \\
ADE20K 1/8      & 2000 & 1008 & 2  & 990 & $0.70$       & $16.59\to16.62$ \\
\bottomrule
\end{tabular}
\end{table*}

Table~\ref{tab:perimage} counts all validation images using saved whole-image predictions.
On Pascal 1/16, PixCon has fewer wrong pixels on 935 images and more on 512. On 1/8 the
counts are 753 and 680. The sign test conditions on this one pair of selected checkpoints;
it does not test variation across training seeds. Per-pixel error and dataset mIoU weight
classes differently, which can explain why the small ADE20K mIoU increase coexists with a slightly
higher pixel-error rate. Cityscapes diagnostics use whole-image inference, whereas
Table~\ref{tab:cityscapes} uses sliding windows; their numerical outcomes are not interchangeable.

The supplement includes selected high-gain examples alongside the full win/loss counts,
three-seed boundary/interior errors, and shared-feature geometry. The boundary and geometry
analyses and the Pascal teacher diagnostic use a separate seed-0 baseline checkpoint ($87.32$
rather than $87.40$).
They are post-hoc descriptions of trained models, not isolated tests of the correctness gate.

\subsection{Control: Clean vs. Predicted Contrastive Labels}
\label{sec:ablation}

Table~\ref{tab:bankablation} compares PixCon with a confidence control. The confidence control uses the same labeled images, projection head, queues and loss as
PixCon. It retains non-void pixels where the \emph{student's} maximum softmax score is at least
0.95 and uses the student argmax as the contrastive class. That class labels both the current
anchor and its future bank entry. PixCon instead requires prediction--ground-truth agreement
and uses the ground truth. The control therefore changes eligibility and anchor/key labels
together. It is not an isolated bank-only intervention or a reproduction of ReCo or
U\textsuperscript{2}PL, which use unlabeled data differently.

\begin{table}[t]
\centering
\caption{\textbf{Two contrastive labeling rules on Pascal.} Best validation EMA mIoU,
seed 0, 16 labeled and 16 unlabeled images per iteration. Both branches draw contrastive
features from labeled images. The predicted-label control thresholds student confidence and
assigns student labels to anchors and keys; PixCon uses ground-truth labels and a correctness
gate. Differences of at most $0.02$ points do not establish equivalence.}
\label{tab:bankablation}
\small
\begin{tabular}{lcc}
\toprule
Contrastive branch & 1/16 & 1/8 \\
\midrule
None (consistency baseline) & $84.47$ & $87.40$ \\
Student-predicted labels, conf.\ $\geq0.95$ & $85.54$ & $87.58$ \\
Ground-truth labels, correct (PixCon) & $85.53$ & $87.60$ \\
\bottomrule
\end{tabular}
\end{table}

Both branch variants score above the no-branch baseline on this seed, by about 1.1 points
at 1/16 and 0.2 at 1/8. Their near-identical scores provide no evidence that the clean rule
accounts for the branch's accuracy gain. PixCon does offer a simpler label-source property:
its contrastive targets cannot inherit pseudo-label assignment errors. The control still uses
ground truth for the supervised loss, so it does not remove annotation access from training.

Neither the confidence bank's training contamination nor its similarity-weighted false-positive
mass was recorded. The validation teacher diagnostics in section~\ref{sec:cleanpos} cannot
supply those missing quantities. A labeled-only branch without the correctness gate was not
run, and no predicted-label contrastive arm was run on ADE20K. Thus, the separate benefit of
correctness filtering and its behavior under noisier contrastive targets remain unresolved.

\section{Conclusion}
\label{sec:conclusion}

A small labeled-pixel contrastive branch can accompany a strong foundation-model consistency
baseline without changing inference. In our fixed-split DINOv2-B study, PixCon raises the Pascal
1/8 mean by $0.89$ mIoU points across three positive paired differences. Other settings show
smaller or mixed changes. The predicted-label control reaches similar Pascal scores, which
suggests that the practical effect comes from adding the branch rather than from its
annotation-consistent bank labels; one seed cannot establish this.
The exact gradient analysis further distinguishes label-count contamination from
similarity-weighted influence. Together, the method and evidence provide a simple reference
configuration with explicit costs and a clear account of what the experiments establish.

\section{Limitations}
\label{sec:limitations}
Three training seeds on one fixed label split are limited evidence: at Pascal 1/8 the mean
paired gain is $0.89\pm0.85$ points (sample std; paired $t(2)=1.82$, two-sided $p=0.21$).
The observed spread reduction is not established statistically. Other datasets have one seed,
and all cells use one backbone and one local host implementation. Best-checkpoint selection
and sensitivity tuning use validation data; no independent test result is reported.
There is no matched supervised-only, labeled-only contrastive, or prior contrastive-method
baseline. The confidence control changes anchor and key labels together, and the teacher
error diagnostic does not measure training-bank noise. Annotation errors and stale features
remain possible in a labeled bank. Measured iteration overhead is $15.3$--$19.7\%$; peak memory
and equal-wall-clock comparisons are unavailable. These limits prevent a general accuracy,
variance-reduction or state-of-the-art claim.

\ifdefined\llmstatement\input{llm_statement}\fi

{
    \small
    \bibliographystyle{ieeenat_fullname}
    \bibliography{references}
}

\ifincludesupp
\clearpage
\appendix
{\centering\normalfont\Large\bfseries Supplementary Material\par}
\vspace{1em}
\section{Reading Guide}
This supplement provides the derivation, implementation details, saved-run provenance,
per-class results and post-hoc diagnostics supporting the main paper. Main-paper tables
report best validation EMA checkpoints. Single-seed controls, teacher diagnostics and
post-hoc feature analyses answer different questions and are identified separately below.
No additional training is used in these analyses.

\section{Exact Gradient and Its Scope}
\label{app:gradient}
For an anchor $z_a$, define $Z_P=\sum_{p\in P}\exp(z_a^\top p/\eta)$ and
$Z_B=\sum_{b\in B}\exp(z_a^\top b/\eta)$. The implemented loss is
$\ell_a=-\log Z_P+\log Z_B$. Bank entries are constants during this differentiation, giving
\begin{align}
\nabla_{z_a}\log Z_P &= \eta^{-1}\sum_{p\in P}q_pp,\\
\nabla_{z_a}\log Z_B &= \eta^{-1}\sum_{b\in B}r_bb.
\end{align}
Subtracting yields Eq.~\eqref{eq:exact-gradient} in the main paper. This derivation applies
to the log of the positive sum. For the alternative loss that averages per-positive log
probabilities, the positive mean is uniform instead; the two objectives should not be conflated.

\paragraph{Mislabeled-positive mass.}
Partition $P=T\cup F$, relative to a correctly labeled anchor, and let
$Z_T$ and $Z_F$ be the corresponding exponential sums. Then
$\alpha=Z_F/(Z_T+Z_F)$ and
$\mu_P=(1-\alpha)\mu_T+\alpha\mu_F$. Since weighted means of unit vectors have norm
at most one,
\begin{equation}
\|\mu_P-\mu_T\|=\alpha\|\mu_F-\mu_T\|\leq2\alpha.
\end{equation}
Also $Z_F/Z_T=(|F|/|T|)(\bar e_F/\bar e_T)$, proving
Eq.~\eqref{eq:mass-ratio}. Unit norms and a shared temperature do not make $\bar e_F$
and $\bar e_T$ comparable. At $\eta=0.1$, a similarity difference of 0.5 multiplies the
per-entry exponential weight by $e^5$, about 148. Consequently, even a small count fraction
can receive substantial softmax mass. Opposing vectors can also cancel, so a coefficient
ratio is not a ratio of gradient magnitudes.

\paragraph{Normalization and the denominator.}
If the projection head outputs $v_a$ and $z_a=v_a/\|v_a\|$, then
\begin{equation}
\nabla_{v_a}\ell_a
=\frac{I-z_az_a^\top}{\|v_a\|}\nabla_{z_a}\ell_a.
\end{equation}
Changing the bank can alter both $\mu_P$ and $\mu_B$. The bound above compares positive
means for fixed embeddings; it does not bound the full change caused by replacing entries,
a trajectory of parameter updates, or final mIoU. The predicted-label control can also
mislabel anchors, whereas the partition above assumes the anchor's class is correct.

\paragraph{Two conditions, two roles.}
Ground-truth class assignments eliminate pseudo-label assignment error relative to dataset
annotations. Prediction--label agreement further restricts eligibility, selecting features
the student currently classifies correctly. A labeled-only bank would have the same
label-source property, but that control was not run. Correct annotations can still correspond
to features that overlap in embedding space, and detached entries can become stale. The
analysis therefore establishes a property of the objective, not the correctness gate's
accuracy benefit.

\section{Implementation Details}
\label{app:implementation}

\paragraph{Memory and compute.}
The branch adds one projection-head forward on the labeled batch, an InfoNCE over $\le\!1024$
anchors against the bank, and the enqueue (no gradient through enqueued features). The head is a
$1{\times}1$ convolution ($256{\to}256$, no bias), batch-norm, ReLU and a second $1{\times}1$
convolution ($256{\to}256$, with bias): $131{,}840$ parameters, about $0.13\%$ of the segmentor,
and it is external to the segmentor in the implementation used for the reported runs.
The inference state has the same architecture as our baseline; its learned weights differ. The bank is $K\!\cdot\!N\!\cdot\!D$ floats in fp32: $21\!\cdot\!256\!\cdot\!256$
$=1.38$M floats ($5.5$~MB) on Pascal, $5.0$~MB on Cityscapes and $39.3$~MB on ADE20K. Nothing from
the branch is used at inference (section~\ref{app:cost} gives the training-time overhead).

\subsection{Hyperparameters}

\begin{table}[h]
\centering
\caption{Hyperparameter settings for each dataset. The settings below are those of our local implementation.
The contrastive settings are fixed across datasets after Pascal sensitivity runs.}
\label{tab:hyperparams}
\small
\resizebox{\linewidth}{!}{%
\begin{tabular}{lccc}
\toprule
Hyperparameter & PASCAL & Cityscapes & ADE20K \\
\midrule
Backbone & DINOv2-Base & DINOv2-Base & DINOv2-Base \\
Decoder & DPT-lite & DPT-lite & DPT-lite \\
Crop size & $518{\times}518$ & $686{\times}686$ & $518{\times}518$ \\
Labeled + unlabeled batch & $16+16$ & $16+16$ & $16+16$ \\
Backbone LR & $5{\times}10^{-6}$ & $5{\times}10^{-6}$ & $5{\times}10^{-6}$ \\
Decoder LR & $2{\times}10^{-4}$ & $2{\times}10^{-4}$ & $2{\times}10^{-4}$ \\
Weight decay & 0.01 & 0.01 & 0.01 \\
Maximum epochs & 60 & 120 & 60 \\
Optimizer & AdamW & AdamW & AdamW \\
LR schedule & poly $p{=}0.9$ & poly $p{=}0.9$ & poly $p{=}0.9$ \\
Conf.\ threshold $\tau$ & 0.95 & 0.95 & 0.95 \\
Labeled oversampling & Yes & Yes & Yes \\
\midrule
\multicolumn{4}{c}{\emph{Clean-positive contrastive branch}} \\
\midrule
Weight $\lambda_\mathrm{pix}$ & 0.1 & 0.1 & 0.1 \\
Temperature $\eta$ & 0.1 & 0.1 & 0.1 \\
Projection dim $D$ & 256 & 256 & 256 \\
Bank size per class $N$ & 256 & 256 & 256 \\
Anchors per class (cap) $m$ & 64 & 64 & 64 \\
Max anchors per iter & 1024 & 1024 & 1024 \\
EMA decay $\gamma_i$ & \multicolumn{3}{c}{$\min(1-1/(i{+}1),\;0.996)$} \\
\bottomrule
\end{tabular}}
\end{table}

\paragraph{Architecture and interpolation.}
The encoder exposes transformer blocks 3, 6, 9 and 12 (one-based indexing). Each feature map
is projected to 256 channels and resampled into a four-level pyramid, then fused from coarse
to fine. The segmentation head uses a $3\times3$ convolution, batch normalization, ReLU,
dropout and a final $1\times1$ class projection. Bilinear interpolation produces image-sized
logits. For the contrastive gate, labels are resized to the fused-feature resolution with
nearest-neighbor interpolation, while student logits use bilinear interpolation with
\texttt{align\_corners=True} before argmax. The projection head has no confidence input.

\paragraph{Checkpoint selection and implementation provenance.}
The reported experiments use the archived research training implementation. The later
standalone reference implementation integrates the projection head into the model and is
not the source of the reported training runs. The research implementation keeps the
projection head and FIFO state outside the saved segmentor checkpoint; saved segmentor
weights support inference, but do not contain all auxiliary state needed for an exact
training restart. Early stopping uses 20 epochs without improvement in validation EMA mIoU;
all runs are selected by the same rule. The training entry point seeds PyTorch and NumPy, but does not seed Python's
random module used by augmentation. Runs are therefore not bitwise reproducible from the
reported seed. A common seed also does not imply identical sampling sequences once the
contrastive branch consumes random numbers.

\subsection{Data Augmentation}

For labeled images: random scaling (0.5--2.0), random cropping, horizontal flipping.
For unlabeled images: additionally color jitter (probability 0.8), random grayscale (0.2),
Gaussian blur ($\sigma \in [0.1, 2.0]$, probability 0.5), and CutMix.
Spatial transforms keep image, label and confidence maps aligned.

\subsection{Computational Cost}
\label{app:cost}

Table~\ref{tab:cost} gives the training-time overhead of the branch, measured from the timestamped
training logs of paired runs that share a container, a crop size, a batch size and an epoch budget:
for each epoch we divide the time between its first and last logged iteration by the number of
iterations between them, and report the median over epochs (over all epochs with at least two valid timestamped iteration records). Wall-clock totals are not a fair comparison
because patience-based early stopping ends each run at a different epoch. The logs do not reliably identify the GPU model for each run or record peak memory;
neither is inferred from later hardware availability. These timing measurements are specific
to the recorded environment and exclude validation and checkpoint I/O.

\begin{table}[h]
\centering
\caption{\textbf{Training overhead of the clean-positive branch.} Median seconds per iteration
over epochs, from the training logs of paired UniMatch~V2 and PixCon runs in the same container
(same crop, batch $16$, same epoch budget), and the resulting overhead. The ADE20K pair ran across
several restarts on a shared host and is omitted as unreliable.}
\label{tab:cost}
\small
\resizebox{\linewidth}{!}{%
\begin{tabular}{lccc}
\toprule
Cell & UniMatch~V2 s/it & PixCon s/it & overhead \\
\midrule
Pascal 1/4      & $0.678$ & $0.782$ & $+15.3\%$ \\
Cityscapes 1/16 & $1.226$ & $1.466$ & $+19.5\%$ \\
Cityscapes 1/8  & $1.220$ & $1.461$ & $+19.7\%$ \\
\bottomrule
\end{tabular}}
\end{table}

\section{Pseudocode}
\label{app:pseudocode}

The PixCon training step is summarized below. The supervised forward extracts the fused
decoder feature (which our segmentor already exposes via a kwarg); the projection head
runs once per iteration over the labeled batch only. The clean-positive bank is updated
\emph{after} loss computation, so anchors from the current iteration do not appear in
their own positives.

\begin{figure*}[t]
\footnotesize
\begin{verbatim}
# logits_l: [B, K, H, W]  fused_l: [B, C, h', w']  mask_l: [B, H, W]
loss_x = CE(logits_l, mask_l)                 # supervised CE

z = head(fused_l); z = normalize(z, dim=1)    # [B, D, h', w']
gt_lo  = interpolate(mask_l,  size=(h', w'), 'nearest')
pred_lo= interpolate(logits_l, size=(h', w'), 'bilinear').argmax(1)
valid  = (gt_lo != 255) & (pred_lo == gt_lo)  # clean-positive filter
anchors, labels = balanced_sample(z[valid], gt_lo[valid], cap_per_class=m)
loss_pix = infonce(anchors, labels, bank, tau=eta)
bank.enqueue(anchors.detach(), labels)

# Unlabeled streams (UniMatch V2): two strong-aug views + complementary
# channel dropout + fixed conf threshold 0.95
loss_u = unimatch_v2_consistency(...)

loss = (loss_x + loss_u) / 2 + lambda_pix * loss_pix
\end{verbatim}
\caption{\textbf{PixCon training step (one iteration).} The clean-positive filter
(\texttt{valid}) keeps only labeled pixels the student already classifies correctly; the bank
is updated \emph{after} the loss so anchors never appear in their own positive set.}
\label{alg:step}
\end{figure*}

\section{Full Per-Class Results}
\label{app:perclass}

Figure~\ref{fig:iou_improvement} and Table~\ref{tab:perclass_pascal} use the per-class IoUs
stored with the best EMA results of all three Pascal seeds. They include gains and regressions.
Classes are ordered by baseline IoU, which measures difficulty here, not class frequency.
These results do not by themselves identify a rare-class mechanism.

\begin{table}[h]
\centering
\caption{\textbf{Per-class Pascal validation IoU (\%).} Mean over three training seeds
on one fixed split, at each run's best EMA checkpoint. U: consistency baseline; P: PixCon.
Differences are computed from unrounded values. Class-wise uncertainty is shown in
Fig.~\ref{fig:iou_improvement}; neither the individual differences nor the aggregate means
establish statistical significance.}
\label{tab:perclass_pascal}
\small
\resizebox{\linewidth}{!}{%
\begin{tabular}{lccc@{\quad}ccc}
\toprule
 & \multicolumn{3}{c}{1/16} & \multicolumn{3}{c}{1/8} \\
\cmidrule(lr){2-4}\cmidrule(lr){5-7}
Class & U & P & $\Delta$ & U & P & $\Delta$ \\
\midrule
chair & 46.6 & 49.4 & $+2.8$ & 62.8 & 65.3 & $+2.4$ \\
sofa & 54.9 & 59.5 & $+4.6$ & 71.5 & 74.4 & $+2.9$ \\
diningtable & 61.1 & 60.1 & $-1.0$ & 75.1 & 76.1 & $+1.0$ \\
pottedplant & 71.4 & 73.2 & $+1.9$ & 64.2 & 65.3 & $+1.1$ \\
tvmonitor & 72.6 & 68.5 & $-4.1$ & 69.8 & 73.7 & $+3.9$ \\
bicycle & 81.3 & 82.3 & $+1.0$ & 81.9 & 82.4 & $+0.5$ \\
bottle & 83.7 & 84.6 & $+0.8$ & 84.6 & 84.5 & $-0.2$ \\
boat & 85.8 & 86.1 & $+0.3$ & 82.7 & 83.1 & $+0.4$ \\
car & 90.3 & 90.6 & $+0.3$ & 90.8 & 92.6 & $+1.7$ \\
person & 90.7 & 91.2 & $+0.6$ & 93.0 & 93.9 & $+0.9$ \\
aeroplane & 91.3 & 92.9 & $+1.5$ & 94.1 & 95.7 & $+1.6$ \\
motorbike & 91.6 & 90.6 & $-1.1$ & 91.8 & 93.2 & $+1.5$ \\
train & 94.2 & 94.1 & $-0.1$ & 93.6 & 94.2 & $+0.7$ \\
horse & 94.7 & 94.8 & $+0.1$ & 96.4 & 96.2 & $-0.2$ \\
sheep & 94.7 & 95.6 & $+0.9$ & 96.4 & 96.6 & $+0.1$ \\
bird & 94.8 & 94.5 & $-0.2$ & 96.7 & 95.7 & $-1.0$ \\
dog & 94.8 & 96.2 & $+1.4$ & 96.1 & 96.5 & $+0.4$ \\
bus & 95.3 & 93.5 & $-1.9$ & 95.3 & 95.7 & $+0.4$ \\
background & 95.5 & 95.7 & $+0.2$ & 96.8 & 96.9 & $+0.1$ \\
cat & 95.7 & 95.8 & $+0.1$ & 95.9 & 96.4 & $+0.5$ \\
cow & 96.5 & 96.8 & $+0.3$ & 97.7 & 97.7 & $+0.0$ \\
\midrule
\textbf{mean} & 84.65 & 85.05 & $+0.40$ & 87.01 & 87.90 & $+0.89$ \\
\bottomrule
\end{tabular}}
\end{table}

\begin{figure*}[t]
\centering
\includegraphics[width=0.92\linewidth,height=0.42\textheight,keepaspectratio]{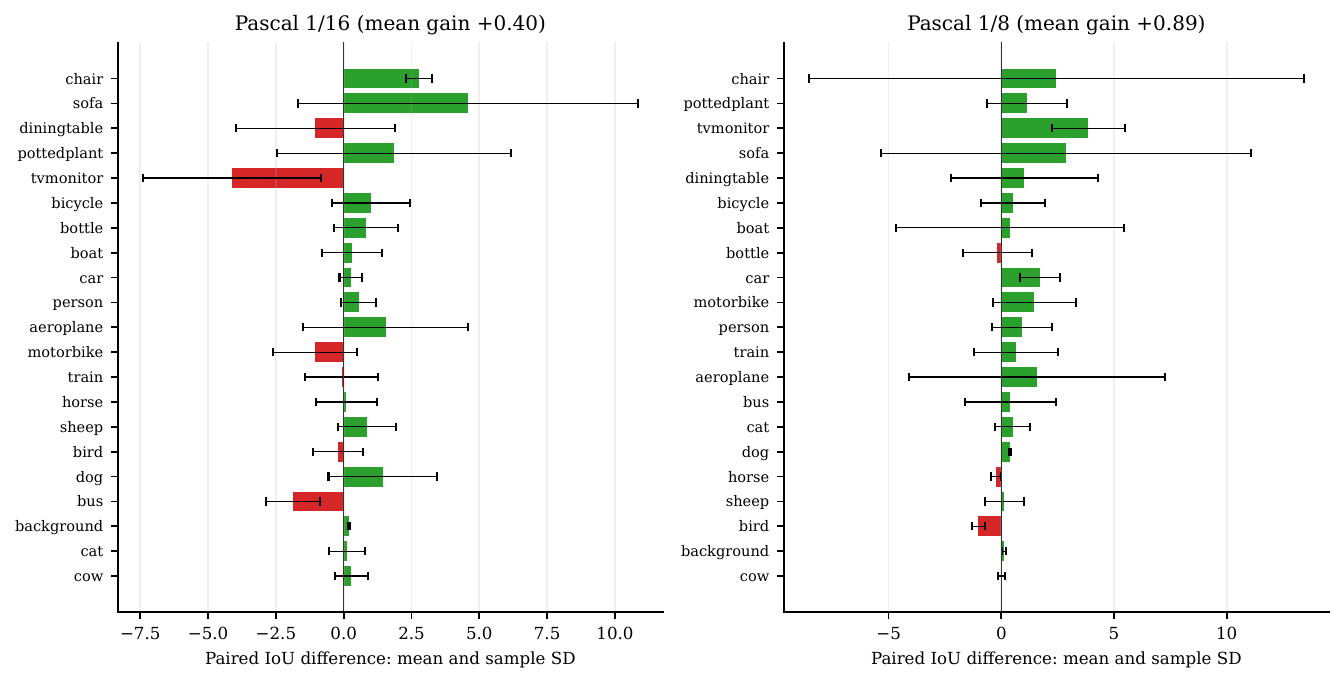}
\caption{\textbf{Per-class changes, including regressions.} Difference in best-EMA
validation IoU between PixCon and our consistency baseline. Bars show the mean of three
paired differences and error bars show their sample standard deviation. Classes are ordered
by baseline IoU, not frequency. The aggregate changes are $+0.40$ at 1/16 and $+0.89$ at 1/8.
Large error bars indicate substantial seed dependence.}
\label{fig:iou_improvement}
\end{figure*}

\section{Boundary/Interior Error, Contamination, and Bank Coverage}
\label{app:newanalyses}

These post-hoc analyses use saved DINOv2-B checkpoints. For the Pascal boundary, teacher-error
and feature-geometry diagnostics, seed 0 uses a separate baseline reproduction ($87.32$ at
epoch 28), rather than the main-table run ($87.40$ at epoch 42). The other five checkpoints
in the three-seed boundary/geometry comparison match the main results. These diagnostics
describe those saved models and do not isolate the correctness gate. The per-image table and
qualitative selection form a separate saved analysis; Cityscapes uses whole-image inference
for those diagnostics, unlike the sliding-window main results.

\subsection{Boundary vs.\ Interior Error}
\label{app:boundary}
Table~\ref{tab:boundary} gives the full breakdown behind the claim in section~\ref{sec:qualitative}.
For each of the $1449$ Pascal~VOC val images we run whole-image EMA-teacher inference and
partition valid pixels into a ground-truth \emph{boundary band} and the remaining interior.
The band starts with pixels adjacent to a label transition, including object/void transitions,
and expands by $w-1$ four-connected binary dilations. Void pixels are then excluded. We report
the per-pixel error rate (\%, lower better) in each region, averaged over the three seeds. At each band width, the mean absolute error reduction is larger at boundaries.
Boundary errors also start much higher, and interiors contain most valid pixels. This is a
pixel-error breakdown, not a boundary-IoU evaluation or evidence of a boundary-specific mechanism.

\begin{table}[h]
\centering
\caption{\textbf{Post-hoc error by distance from a label boundary.} Pascal~VOC 1/8
val error rate (\%, lower better), DINOv2-B EMA teachers, mean over 3 seeds. Boundary band = label-transition pixels plus $w-1$ four-connected dilations; interior = the rest. $\Delta$ = UM2 repro $-$ PixCon
(positive means PixCon better). The boundary band is ${\approx}5\%$ of valid pixels at $w{=}3$;
sample standard deviations and paired values are in the saved analysis; the table shows means.}
\label{tab:boundary}
\small
\resizebox{\linewidth}{!}{%
\begin{tabular}{lcccc}
\toprule
band $w$ & boundary UM2$\to$PixCon & $\Delta$bd & interior UM2$\to$PixCon & $\Delta$in \\
\midrule
$1$~px & $15.78 \to 15.48$ & $+0.30$ & $2.58 \to 2.41$ & $+0.18$ \\
$3$~px & $11.97 \to 11.62$ & $+0.35$ & $2.34 \to 2.17$ & $+0.17$ \\
$5$~px & $10.04 \to \phantom{0}9.68$ & $+0.35$ & $2.21 \to 2.04$ & $+0.17$ \\
\bottomrule
\end{tabular}}
\end{table}

\subsection{Retained-Pixel Teacher Error Across Datasets}
\label{app:rhof}
Table~\ref{tab:rhof} reports $e_{\mathrm{ret}}$, the pixel-pooled error rate of the baseline
EMA teacher among non-void validation pixels with confidence at least 0.95. It is measured
at the baseline's best-EMA checkpoint, on unaugmented validation images. Neither per-class FIFO
sampling nor student predictions during training enter this computation. It is therefore
a teacher diagnostic, not a measurement of any method's realized training-bank contamination.
The ADE20K labels are remapped from 1--150 to 0--149, with 0 ignored. The ADE20K inference
check gives 49.09 versus 49.10 stored mIoU, a 0.01-point difference; Pascal matches to two
decimals. The error rate is about six times higher on ADE20K for these two checkpoints.

\begin{table}[h]
\centering
\caption{\textbf{Retained-pixel error of the baseline EMA teacher.} Whole-image validation
inference at confidence threshold 0.95. Retention and accuracy are pooled over valid pixels.
$e_{\mathrm{ret}}$ is not the training-bank quantity $\rho_F$ in the main-paper derivation.}
\label{tab:rhof}
\small
\resizebox{\linewidth}{!}{%
\begin{tabular}{lcccc}
\toprule
Dataset & mIoU (sanity) & retention & retained acc & $e_{\mathrm{ret}}$ \\
\midrule
Pascal 1/8 & $87.32$ & $97.6\%$ & $98.2\%$ & $0.018$ \\
ADE20K 1/8 & $49.09$ & $85.7\%$ & $89.4\%$ & $0.106$ \\
\bottomrule
\end{tabular}}
\end{table}

\subsection{Coarse-Grid Eligibility on ADE20K}
\label{app:coverage}
A saved post-hoc diagnostic applies the correctness gate of the PixCon student weights stored in the best-EMA checkpoint to one pass
over the 2,526 labeled ADE20K images. Images are resized to $518\times518$ and predictions
and labels are reduced to a $37\times37$ grid. All 150 classes have at least one eligible
point; 148 classes have at least 256, with minimum 28 and median 2,908 eligible points per
class over the pass. This is an eligibility count at the ViT token grid, not a replay of
the training queue: training projects fused decoder features at a higher resolution and uses
augmentation, per-class sampling and FIFO replacement. The count cannot establish actual
queue occupancy, turnover or early-training coverage, nor rule out coverage as a source of
class-specific failures. Partially filled queues remain usable; only anchors lacking a
same-class entry are excluded from the contrastive loss.

\subsection{Feature-Space Geometry of the Shared Embedding}
\label{app:geometry}
We describe the shared decoder representation on the saved checkpoints. We compare the one representation the two
models share and that actually feeds the classifier: the fused DPT decoder feature (the PixCon
projection head is auxiliary, is discarded at inference, and has no UniMatch~V2 counterpart, so it
is not a fair axis of comparison). For each of the $1449$ Pascal val images we take the fused
feature, $\ell_2$-normalize it per pixel, and group pixels by ground-truth label. From the
per-class sums we read off, in one pass and exactly, (i) \emph{intra-class compactness}, the mean
cosine of a class's pixels to their unit class centroid (the mean resultant length; higher is
tighter), (ii) \emph{inter-class cosine}, the mean and max pairwise cosine between class centroids
(lower is better separated), and (iii) their difference, the \emph{margin}. We average over the
$20$ foreground classes and report the mean over the three seeds (Table~\ref{tab:geometry}).

The mean margin increases by 0.014, with positive differences in all three saved pairs.
Mean inter-class cosine decreases in two pairs; intra-class compactness changes by only 0.003.
These are descriptive feature statistics. They neither establish that the gate caused the
change nor show a general reduction in seed variability. The saved script resizes the unpadded
ground-truth map directly to the feature grid of a patch-padded image. This introduces a
spatial-alignment approximation near transitions; the reported geometry is retained as an
exploratory diagnostic, not a precise measure of boundary feature structure. The full-dimensional cosine statistics are the reported readout; independently fitted
two-dimensional visualizations are not used as evidence.

\begin{table}[h]
\centering
\caption{\textbf{Shared decoder-feature statistics of saved checkpoints.} Geometry
of the shared fused decoder feature on Pascal~VOC 1/8 (val, $20$ foreground classes), DINOv2-B EMA
teachers, mean$\pm$sample std over 3 seeds. Compactness $=$ mean cosine to class centroid (higher tighter);
inter $=$ mean/max centroid-pair cosine (lower better); margin $=$ compactness $-$ mean inter.
$\Delta =$ PixCon $-$ baseline; the sign is not reversed for lower-is-better quantities.}
\label{tab:geometry}
\small
\resizebox{\linewidth}{!}{%
\begin{tabular}{lccc}
\toprule
Quantity & UM2 repro & PixCon & $\Delta$ \\
\midrule
Intra-class compactness ($\uparrow$) & $0.911{\pm}0.012$ & $0.914{\pm}0.006$ & $+0.003$ \\
Mean inter-class cosine ($\downarrow$) & $0.406{\pm}0.003$ & $0.395{\pm}0.012$ & $-0.011$ \\
Max inter-class cosine ($\downarrow$)  & $0.539{\pm}0.012$ & $0.524{\pm}0.028$ & $-0.015$ \\
Margin ($\uparrow$) & $0.505{\pm}0.010$ & $0.519{\pm}0.013$ & $+0.014$ \\
\bottomrule
\end{tabular}}
\end{table}

\section{Training Dynamics, Per-Class Breakdown, and Sensitivity}

This section collects results deferred from the main paper for space.

\subsection{Training Dynamics}
The per-epoch training curves (Fig.~\ref{fig:curves}, main paper) show the 3-seed mean PixCon curve
leading UniMatch~V2 in $41/41$ all-live epochs at 1/8 and in $26/28$ at 1/16 (one tie at the logged precision). Runs differ in length
because patience-based early stopping halts each once its EMA mIoU plateaus; the three-seed mean curves stop at the end of the common window, while individual curves continue.

\paragraph{Paired-seed uncertainty.}
From unrounded best-EMA results, Pascal 1/8 has mean paired gain $0.889$ and sample standard
deviation $0.845$ points; the two-sided paired $t$ test gives $p=0.210$. The corresponding
1/16 values are $0.396$, $0.606$ and $p=0.374$. With three pairs, the normality assumption
cannot be assessed and uncertainty is wide. The per-image sign tests condition on one
checkpoint pair and should not be substituted for these across-run comparisons. Repeated
validation epochs are correlated observations of the same training runs.

\paragraph{Peak epochs.}
At 1/16, the baseline peaks at epochs 8/11/28 and PixCon at 39/39/19. At 1/8, the respective
peaks are 42/21/40 and 22/41/34. These records show different training trajectories and
stopping times; they do not isolate a mechanism that prolongs useful learning.

\subsection{Per-Class Analysis}
\label{sec:perclass_analysis}

At 1/8 the largest mean gains include tvmonitor, sofa and chair, while bird regresses.
At 1/16 sofa and chair improve, but tvmonitor and bus regress. Most classes have mixed-sign
paired differences across seeds. Without class-frequency measurements or an isolated
intervention, these observations do not establish or rule out a rare-class mechanism.
They describe where the aggregate change comes from in the evaluated runs.

\subsection{Hyperparameter Sensitivity}
\label{sec:sensitivity}

\begin{table}[t]
\centering
\caption{\textbf{Single-seed sensitivity at batch 4.} Sensitivity of best EMA
mIoU (\%) on Pascal~VOC 1/8 (DINOv2-Base, batch~$4$, seed~$0$) to the contrastive weight
$\lambda_\mathrm{pix}$, temperature $\eta$, and per-class bank size $N$. Each block varies one
hyperparameter with the others at their default (\underline{underlined}: $\lambda_\mathrm{pix}
{=}0.1$, $\eta{=}0.1$, $N{=}256$); the three blocks share the same default run ($86.95$, peak
epoch~$15$). Each tested alternative scores lower than the shared default in this single-seed sweep.}
\label{tab:sensitivity}
\small
\begin{tabular}{llccc}
\toprule
 & setting & variant 1 & \underline{default} & variant 2 \\
\midrule
$\lambda_\mathrm{pix}$ & value     & 0.3   & \underline{0.1} & 0.5 \\
                       & mIoU      & $85.55$ & $\mathbf{86.95}$ & $86.23$ \\
                       & peak ep   & 5     & 15  & 7 \\
\midrule
$\eta$ & value     & 0.07  & \underline{0.1} & 0.2 \\
       & mIoU      & $85.97$ & $\mathbf{86.95}$ & $85.81$ \\
       & peak ep   & 5     & 15  & 2 \\
\midrule
$N$ & value     & 128   & \underline{256} & 512 \\
    & mIoU      & $85.48$ & $\mathbf{86.95}$ & $85.91$ \\
    & peak ep   & 14    & 15  & 3 \\
\bottomrule
\end{tabular}
\end{table}

Table~\ref{tab:sensitivity} reports the saved batch-4, seed-0 sweeps. The default is the
best of these tested settings; no setting below $\lambda_{\mathrm{pix}}=0.1$ was evaluated.
The lower scores and earlier peaks at other settings do not establish instability or
statistical sensitivity without repeats. Batch size also differs from the main comparison,
so these results cannot quantify robustness of the batch-16 recipe.

\paragraph{Stricter consistency retention: an additional negative control.}
An incomplete archived batch-4, seed-0 Pascal 1/8 run raises the consistency threshold to
$\max(0.95,Q_{0.2}(c))$, where $Q_{0.2}$ is the batch confidence quantile. This excludes at
least the lowest-confidence fifth when that quantile exceeds 0.95; it never lowers the
threshold. The retained log reaches epoch 8 and records a best EMA score of 85.37 at epoch 3.
It does not contain a completed run, so a final comparison with the fixed-threshold run
is unavailable. The intervention concerns reduced coverage, not admitting noisier
pseudo-labels, and does not test the contrastive-bank analysis.

\section{Data, Reproducibility and Scope}
All reported training uses existing segmentation benchmarks and their supplied annotations.
The label-source guarantee is relative to those annotations; it does not verify their
semantic correctness. Post-hoc ADE20K diagnostics apply the required void and class-ID remap.
The Cityscapes qualitative diagnostic uses a materialized validation mirror with mapped
19-class training IDs; it is separate from the official-split training results.

The work evaluates segmentation accuracy and training cost, not deployment safety, fairness
or robustness under distribution shift. Errors on small or rare objects can matter in
applications such as driving even when aggregate mIoU is high. The method is not evaluated
as a safety-critical system. No new human-subject data collection is part of the reported
experiments.

For reproducibility, the saved run records identify the split, seed, configuration, best-EMA
score and selection epoch. The three-seed cells keep the labeled split fixed. The standalone
reference code and inference checkpoints do not by themselves recover all historical
training state. The saved split file lists have no exact image-ID overlap between labeled, unlabeled
and validation sets in the six paired cells. Near-duplicate image overlap and possible overlap with DINOv2's pretraining corpus have not
been independently audited. No claim of pretraining-data independence is made.

\section{Published Cross-Dataset Reference Results}
The following tables retain the published context for the single-seed comparison in the main
paper. Methods with different encoders and protocols are not matched controls for PixCon.
These rows are transcribed from the comparison tables in UniMatch~V2 (cited in the main paper).

\begin{table}[t]
\centering
\caption{\textbf{Cityscapes, validation mIoU (\%).} All our results are seed 0, best EMA,
with crop/window 686. Published reference rows are quoted from UniMatch~V2~\cite{unimatchv2}.
Only the last two rows are protocol-matched. The differences are $+0.03$/$-0.08$ at
1/16 and 1/8; no variance estimate is available. Our 1/4 baseline is unpaired.}
\label{tab:reference_cityscapes}
\small
\setlength{\tabcolsep}{5pt}
\resizebox{\linewidth}{!}{%
\begin{tabular}{llcccc}
\toprule
Method & Encoder & 1/16 & 1/8 & 1/4 & 1/2 \\
\midrule
\multicolumn{6}{l}{\emph{Specialized backbones (prior work):}} \\
AugSeg~\cite{augseg}             & RN-101  & 75.2 & 77.8 & 79.6 & 80.4 \\
UniMatch~\cite{unimatch}         & RN-101  & 76.6 & 77.9 & 79.2 & 79.5 \\
CorrMatch~\cite{corrMatch}       & RN-101  & 77.3 & 78.5 & 79.4 & 80.4 \\
BeyondPixels~\cite{beyondpixels} & RN-101  & 78.5 & 79.2 & 80.9 & 81.3 \\
SemiVL~\cite{semivl}             & CLIP-B  & 77.9 & 79.4 & 80.3 & 80.6 \\
\midrule
\multicolumn{6}{l}{\emph{DINOv2 reference results and our runs:}} \\
UniMatch~V2 (published)~\cite{unimatchv2} & DINOv2-S & 80.6 & 81.9 & 82.4 & 82.6 \\
UniMatch~V2 (published)~\cite{unimatchv2} & DINOv2-B & 83.6 & 84.3 & 84.5 & 85.1 \\
UniMatch~V2 (our repro)                  & DINOv2-B & $\mathit{83.16}$ & $\mathbf{83.96}$ & $83.99$ & -- \\
PixCon (ours)                            & DINOv2-B & $\mathbf{83.20}$ & $\mathit{83.88}$ & -- & -- \\
\bottomrule
\end{tabular}}
\end{table}

\begin{table}[t]
\centering
\caption{\textbf{ADE20K, validation mIoU (\%).} Headers give label counts. Published
reference rows are quoted from UniMatch~V2~\cite{unimatchv2}. Our 1/8 results use seed 0 and
best EMA selection. The $+0.14$-point difference has no repeat-run uncertainty estimate.}
\label{tab:reference_ade20k}
\small
\setlength{\tabcolsep}{4.5pt}
\resizebox{\linewidth}{!}{%
\begin{tabular}{llccc}
\toprule
Method & Encoder & 1/32\,(631) & 1/16\,(1263) & 1/8\,(2526) \\
\midrule
UniMatch~\cite{unimatch} & RN-101  & 28.1 & 31.5 & 34.6 \\
UniMatch~\cite{unimatch} & CLIP-B  & 31.2 & 34.4 & 38.0 \\
SemiVL~\cite{semivl}     & CLIP-B  & 35.1 & 37.2 & 39.4 \\
UniMatch~V2 (published)~\cite{unimatchv2} & DINOv2-S & 38.1 & 40.7 & 44.4 \\
UniMatch~V2 (published)~\cite{unimatchv2} & DINOv2-B & 45.0 & 46.7 & 49.8 \\
\midrule
UniMatch~V2 (our repro)  & DINOv2-B & -- & -- & $\mathit{49.10}$ \\
PixCon (ours)            & DINOv2-B & -- & -- & $\mathbf{49.23}$ \\
\bottomrule
\end{tabular}}
\end{table}

\section{Additional Qualitative Comparisons}
\label{app:qual_full}

The per-image comparison over every validation image is in the main paper
(section~\ref{sec:qualitative}); the figures here illustrate what changes on the images where PixCon
gains most, and they are selected. For each cell we rank validation images by the net error
advantage of PixCon over the UniMatch~V2 reproduction (wrong pixels of the baseline minus wrong
pixels of PixCon, divided by valid pixels; seed~0, EMA teachers). The Pascal rows in
Fig.~\ref{fig:qualitative_full} were then chosen from the $35$ highest-ranked images for class
variety; their ranks lie between $2$ and $35$. The ADE20K rows (Fig.~\ref{fig:qualitative_ade})
rank $1$ to $15$ and the Cityscapes rows (Fig.~\ref{fig:qualitative_cs}) are the six
highest-ranked. The recurring baseline failure on Pascal is a part-level class confusion
(chair-over-person, dog-/sofa-over-cat, pottedplant and tvmonitor mix-ups) or a large spurious
region (objects hallucinated over background, over-extended table and plant masks), which PixCon
collapses to a single coherent, correctly labeled mask. Images at the other end of the ranking
show the reverse, with PixCon adding errors; Table~\ref{tab:perimage} in the main paper counts
both ends.

\begin{figure*}[p]
\centering
\includegraphics[width=\linewidth,height=0.82\textheight,keepaspectratio]{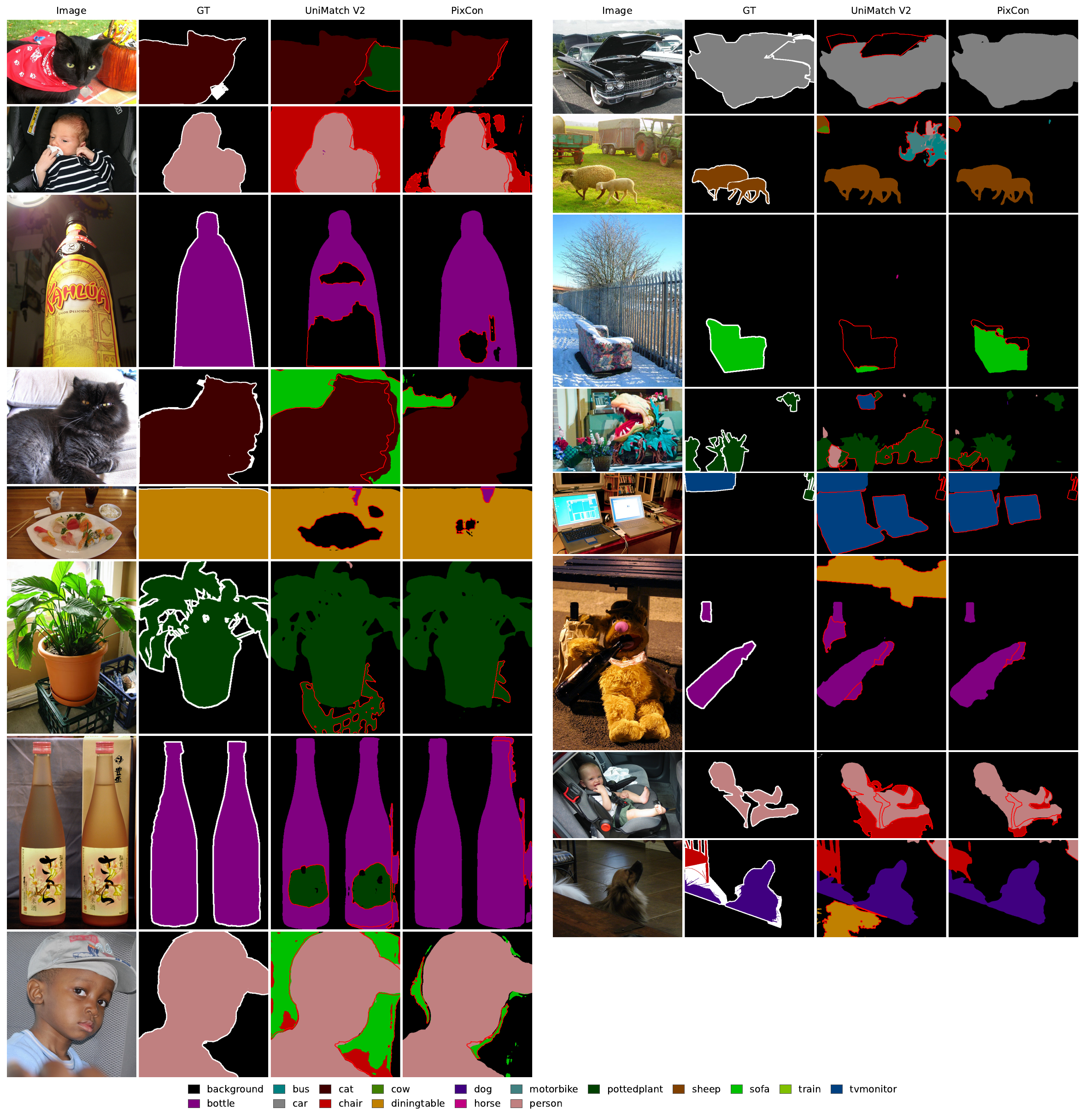}
\caption{\textbf{Sixteen selected PixCon wins on Pascal~VOC 1/8.} Sixteen validation images with
a large per-image error advantage of PixCon over the UniMatch~V2 reproduction (DINOv2-Base,
EMA-teacher predictions, seed~0), chosen for class variety from the $35$ highest-ranked, as two
side-by-side blocks of \emph{input\,$\vert$\,ground truth\,$\vert$\,UniMatch~V2\,$\vert$\,PixCon};
the bottom strip is the shared class palette. \emph{Red contours} mark connected regions where a
prediction disagrees with ground truth (void ignored; components ${<}0.5\%$ of the image
suppressed). The recurring baseline failure is a part-level class confusion on an
already-localized object or a large spurious region; PixCon's panels carry far fewer red contours
on these images. See section~\ref{app:qual_full} for how they were selected.}
\label{fig:qualitative_full}
\end{figure*}

\paragraph{Cityscapes.} Figure~\ref{fig:qualitative_cs} repeats the analysis on Cityscapes 1/16
(19 classes, EMA teachers). The main-table models differ by a small amount here ($+0.03$~mIoU,
Table~\ref{tab:cityscapes} in the main paper) and their predictions are near-identical; even the
largest-per-image-advantage cases differ only in marginal class-boundary corrections, where
PixCon leaves slightly fewer error contours (e.g.\ at building/wall/sidewalk boundaries). These selected examples illustrate small changes; they cannot establish parity across
the validation set.

\begin{figure*}[p]
\centering
\includegraphics[width=\linewidth]{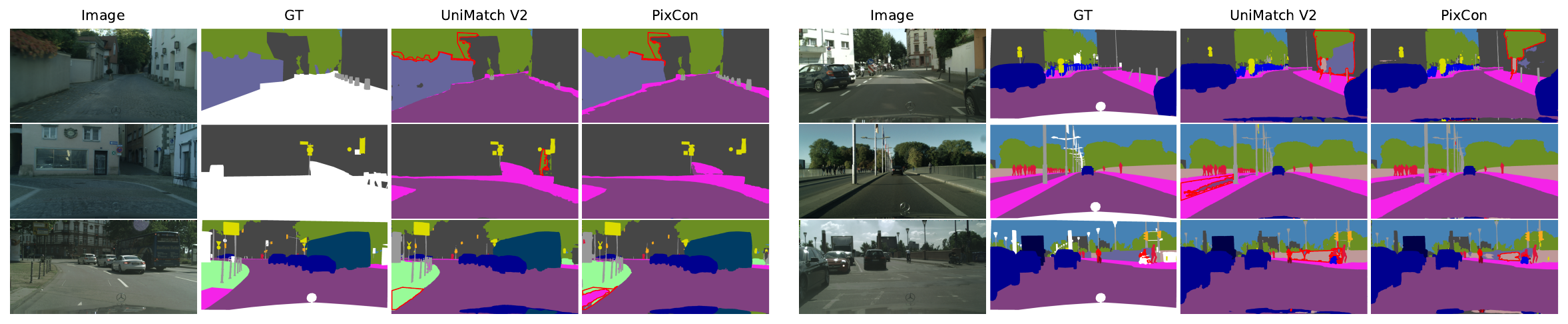}
\caption{\textbf{Cityscapes qualitative (1/16): near-identical, PixCon marginally cleaner.}
Six Cityscapes val images with the largest per-image error advantage of PixCon over the
UniMatch~V2 reproduction (DINOv2-Base, EMA-teacher predictions, seed~0), as two side-by-side
blocks of \emph{input\,$\vert$\,ground truth\,$\vert$\,UniMatch~V2\,$\vert$\,PixCon}; standard
19-class palette, ignore shown white. \emph{Red contours} mark connected regions where a
prediction disagrees with ground truth. On these selected images, the two columns are
nearly identical: PixCon's advantage here is a handful of small boundary corrections (fewer red
contours on a few building/wall/sidewalk regions), not the large class fixes seen on Pascal.}
\label{fig:qualitative_cs}
\end{figure*}

\begin{figure*}[t]
\centering
\includegraphics[width=0.98\linewidth]{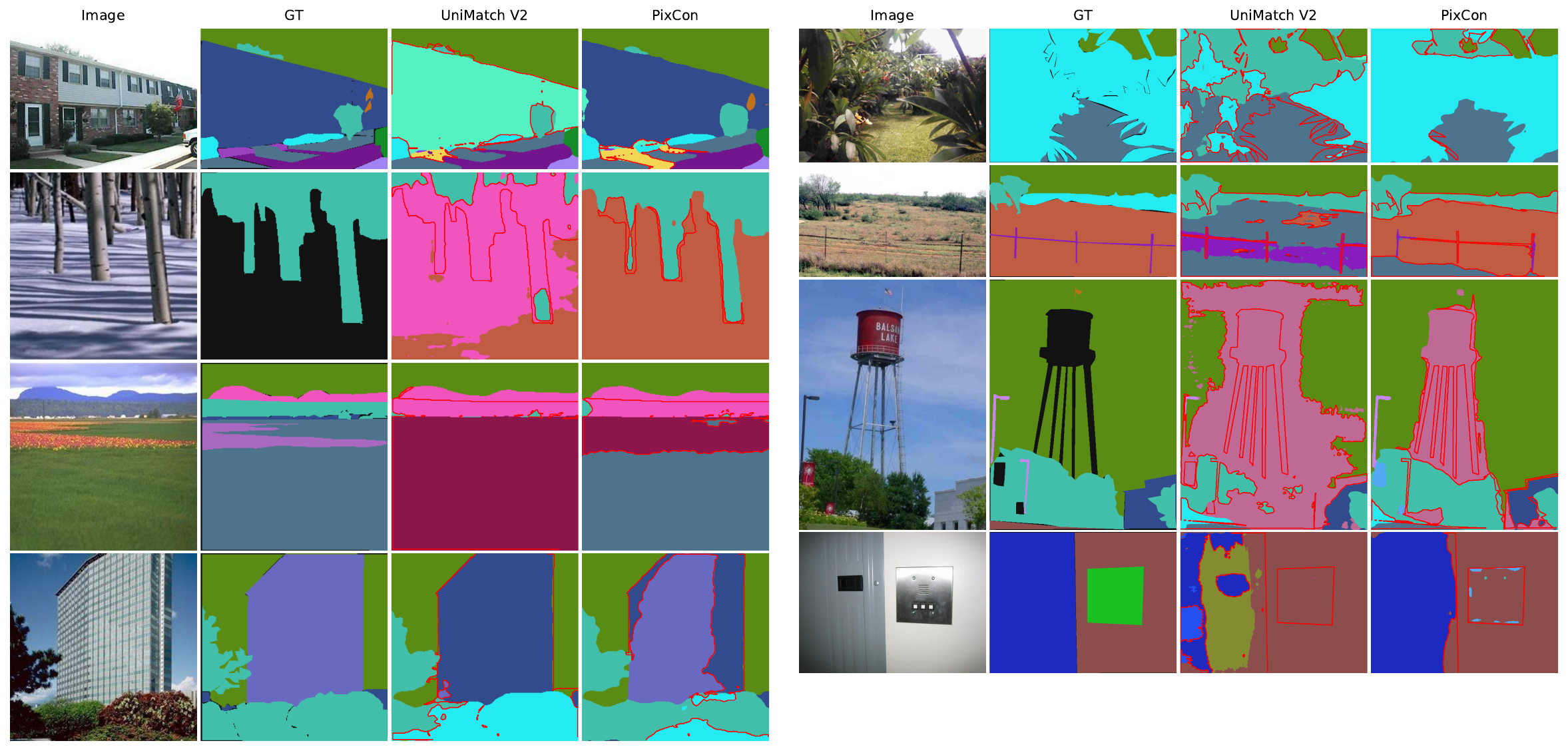}
\caption{\textbf{ADE20K 1/8: eight selected PixCon wins.} Per-image PixCon-vs-UniMatch~V2
comparisons (150-class palette) among the fifteen images with the largest per-image error
advantage, in the same \emph{input\,$\vert$\,GT\,$\vert$\,UniMatch~V2\,$\vert$\,PixCon} format as
Fig.~\ref{fig:qualitative_full}; \emph{red contours} mark disagreement with GT. The main-table models have a small single-seed difference
on ADE20K ($+0.14$~mIoU, Table~\ref{tab:ade20k} in the main paper) and mixed per-image outcomes
($1008$ wins to $990$ losses, Table~\ref{tab:perimage}); these panels show the kind of error that
moves when one does, large scene-parsing mislabellings replaced by the correct class, and an
equally large set of images moves the other way.}
\label{fig:qualitative_ade}
\end{figure*}

\fi

\end{document}